\documentclass[sigconf,nonacm]{acmart}
\setcopyright{none}

\usepackage{tabularx}
\usepackage{array}
\usepackage{float} 
\AtBeginDocument{%
  }

\begin{document}

\title{FieldGuide: A Corpus Map for Token-Efficient Agentic Document Exploration}

\author{Nick Hagar}
\email{nhagar@umn.edu}
\orcid{0000-0001-5110-3737}
\correspondingauthor
\affiliation{%
  \institution{University of Minnesota}
  \city{Minneapolis}
  \state{Minnesota}
  \country{USA}
}

\renewcommand{\shortauthors}{Hagar}

\begin{abstract}
    Generative AI agents can help journalists explore large document collections. But exploratory workflows can require substantial token use and time, especially when the task is broad and the relevant subquestions are not yet known. This paper introduces \textit{FieldGuide}, an agent-facing corpus map that organizes a document collection into a topic hierarchy and exposes tools for corpus orientation, scoped search, and bounded reading. I evaluate FieldGuide on five journalistic document-exploration tasks using 38 Chicago Police Department public-records documents, comparing it with raw-file access and retrieval-augmented generation. Across the five tasks, FieldGuide used 27\% fewer tokens and 41\% less elapsed time than raw-file access; relative to RAG, it used 38\% fewer tokens and 21\% less elapsed time. An accuracy audit identified four minor issues in FieldGuide's responses, comparable to the other conditions. In blind ratings of generated reporting leads, FieldGuide's mean ratings were higher than RAG's and lower than the raw-file condition's. These descriptive results suggest that corpus orientation may reduce runtime token use and elapsed time without severe accuracy failures. They also motivate further study of whether selective access limits broad, cross-document discovery.
\end{abstract}





\maketitle

\section{Introduction}
One promising use case for generative AI (GenAI) agents in journalism is \textit{document exploration}: the sensemaking work through which journalists orient themselves to potentially newsworthy material in a corpus \citep{ettema_epistemology_1984}. Early GenAI newsroom deployments suggest that these systems can substantially reduce the time and labor required to search, summarize, and analyze large document collections \citep{felps_newsrooms_2026,healy_how_2026}. Although caution around hallucinations remains warranted, recent work has also shown that GenAI agents can produce grounded and accurate syntheses of document collections under appropriate constraints \citep{hagar_-premise_2025}.

Yet reducing human labor does not make these systems inexpensive. GenAI agents carry costs of their own---sometimes prohibitively high ones \citep{spencer_how_2026,ona_case_2024}. These costs arise from three sources. First, although LLM inference has become more cost-effective overall, the frontier and long-context models most useful for agentic document exploration remain expensive \citep{sajadieh_ai_2026}. Second, agents consume tokens not only when generating final answers, but also while planning, searching, reading, and revising over many steps of autonomous work \citep{bai_how_2026}. Third, PDFs---a common format for documents of journalistic interest---are often token-dense and require additional processing before agents can reliably access their contents \citep{poznanski_olmocr_2025}.

These challenges point to a need for token-efficient approaches to GenAI-assisted document exploration. Prior work on retrieval-augmented generation (RAG) has largely focused on selecting compact evidence from source documents so that models can ground responses without ingesting an entire corpus \citep{gao_retrieval-augmented_2024}. But journalistic document exploration poses a different kind of problem. Reporting often begins with a broad or evolving information need rather than a precise query. As in other forms of exploratory sensemaking, information needs emerge through iterative cycles of searching, reading, comparison, and reformulation \citep{bates_design_1989}. Journalistic corpora are also heterogeneous, often combining documents that vary in topic, format, and quality \citep{brehmer_overview_2014,wiedemann_newsleak_2018}. A retrieval system optimized for answering a known question may therefore be poorly suited to the upstream task of helping orient a GenAI agent to the contents of a corpus.

To address these challenges, I introduce \textit{FieldGuide}\footnote{\url{https://github.com/NHagar/fieldguide}}, an agent-facing corpus map that exposes a local document collection through a sequence of bounded navigation and evidence operations. At indexing time, FieldGuide extracts and normalizes document text, constructs a hierarchical topic index, and records document, passage, entity, and quality metadata. At runtime, an agent can orient to the corpus, expand topical branches, inspect topic and document metadata, search within an explicit scope, and read bounded text windows. These tools separate navigation from evidence: Compact topic-level objects help the agent choose where to look, while the workflow directs the agent to ground claims in specific passages. FieldGuide is intended to complement retrieval by controlling how broadly the agent searches and how much source text it reads at each step.

I evaluate FieldGuide on five document-exploration prompts using 38 documents from 18 completed Chicago Police Department public-records requests, comparing it with raw-file access and a LanceDB-based RAG workflow. Across the five prompts, the observed FieldGuide runs used 27\% fewer tokens and 41\% less elapsed time than raw-file access; compared with RAG, they used 38\% fewer tokens and 21\% less elapsed time. An accuracy audit identified only minor issues in FieldGuide's responses, similar to the other conditions. In blind ratings of reporting leads' quality and newsworthiness, FieldGuide's observed means were higher than RAG's and lower than the raw-file condition's. Together, these descriptive results suggest that corpus orientation can reduce exploration cost, while also indicating that highly selective access may constrain open-ended editorial discovery.

\section{Related Work}
This work builds on a long line of systems for helping journalists and analysts make sense of large document collections. In visual analytics, systems such as Jigsaw represented documents through entities and relationships to help analysts identify connections across reports \citep{gorg_combining_2013}. Journalism-specific tools adapted similar ideas to the practical demands of newsroom document work. Overview, for example, was designed for systematic analysis of large document collections through clustering, visualization, and tagging \citep{brehmer_overview_2014}. New/s/leak and New/s/leak 2.0 similarly combined linguistic preprocessing with faceted search and visualization to surface salient information in large corpora \citep{yimam_newsleak_2016,wiedemann_newsleak_2018}. Together, these systems treat orientation---seeing the collection's structure before pursuing individual records---as a core part of document analysis. FieldGuide adapts that insight from a visual interface for reporters to a machine-readable set of navigation objects and actions for an agent.

A second strand of work develops LLM-based retrieval and agentic document-query systems. RAG combines LLMs with external evidence retrieved from a corpus, improving accuracy and grounding on knowledge-intensive tasks \citep{lewis_retrieval-augmented_2020,gao_retrieval-augmented_2024}. Rather than asking the model to read everything, a RAG system selects passages likely to answer the user's question. Agentic approaches extend this pattern by allowing models to interleave reasoning with actions such as searching and opening documents \citep{yao_react_2023,asai_self-rag_2023}. This is especially relevant for journalism, where a useful answer may require several rounds of search, comparison, and corroboration. Recent newsroom-oriented work has begun to explore these possibilities by using LLMs to support grounded synthesis over document collections while preserving citations to source material \citep{hagar_-premise_2025}. These systems are strongest when retrieval begins from an articulated query. FieldGuide instead targets an earlier exploratory stage, when searching and reading are also used to refine the information need \citep{bates_design_1989,pirolli_sensemaking_2005}.

Recent work on corpus-level retrieval and summarization attempts to address broad and multi-step questions. GraphRAG constructs graph-based indexes that support query-focused summarization over corpora \citep{larson_graphrag_2024}. RAPTOR similarly builds a hierarchical representation of a corpus, allowing retrieval at multiple levels of abstraction \citep{sarthi_raptor_2024}. Both approaches demonstrate the value of corpus-level abstractions for broad questions. FieldGuide uses a related abstraction for a different control problem: choosing which topical region, document, or passage the agent should inspect next.

Finally, these efforts intersect with work on context and token efficiency in LLM systems. While long context windows allow a model to reason over more of a corpus, those windows also increase cost and latency and may reduce reliability when relevant information is buried in a long input \citep{liu_lost_2024}. Several approaches attempt to address this issue. Selective Context prunes less informative portions of long inputs \citep{li_compressing_2023}, while LLMLingua and LongLLMLingua compress prompts to reduce inference cost and latency \citep{jiang_llmlingua_2023,jiang_longllmlingua_2024}. FieldGuide moves that efficiency intervention upstream: Rather than compressing a task-specific context, it uses a reusable corpus representation to limit how much context the agent constructs.

\section{System Design}
FieldGuide is an access layer between an agent and a local document collection. Its interaction model follows the progression from overview to filtering to details on demand in exploratory interface design \citep{shneiderman_eyes_1996}, as well as accounts of analytic sensemaking in which analysts repeatedly narrow a collection before closely reading and extracting evidence \citep{pirolli_sensemaking_2005}. The system separates two stages: offline index construction and runtime access through seven bounded tools. The completed index remains server-side, so the agent can inspect only the navigation records and source text returned through those tools, rather than loading the underlying index or corpus directly.

\subsection{Index Construction}
At build time, FieldGuide extracts page-aligned text and basic metadata from each source file. It records the extraction method and any warnings, using OCR when native text is unavailable, and divides the resulting canonical text into page-bounded passages. It also records entities and structured values, including people, organizations, locations, email addresses, and monetary amounts.

To provide a reusable view of corpus structure, FieldGuide represents each document with a weighted term profile and recursively groups documents by lexical similarity. The procedure uses deterministic initialization so that the same corpus and settings produce the same hierarchy. In the current implementation, splitting stops after four levels or when a branch contains fewer than six documents. Each topic stores document and page counts, label terms, common entities, neighboring topics, and quality warnings.

\subsection{Runtime Tool Layer}
FieldGuide exposes the index through a Model Context Protocol (MCP) server. MCP allows external systems to present model-invocable tools with structured inputs and outputs.\footnote{\url{https://modelcontextprotocol.io/docs/getting-started/intro}} The runtime interface contains seven operations, organized into three stages.

\textbf{Orientation.} The agent begins with \texttt{orient\_corpus}, which returns corpus counts, source and file-type facets, extraction-quality information, and a limited set of top-level topics. It can then use \texttt{expand\_topic} to inspect child and neighboring topics and \texttt{topic\_card} to view a topic's label terms, entities, representative documents, snippets, and quality warnings. Topic-card responses explicitly identify themselves as navigation aids rather than factual summaries.

\textbf{Triage and scoped retrieval.} The \texttt{search\_within} tool requires an explicit scope, such as a topic subtree, and can further filter by document type, source, or entity. It ranks lexical matches at the document, page, or passage level and diversifies the returned set across documents and sources. The agent can then use \texttt{doc\_card} to inspect document metadata, extraction quality, a page map, candidate passages, and related records before deciding whether to read more closely.

\textbf{Evidence reading.} The \texttt{read\_window} tool returns source text surrounding a selected snippet or character range, while \texttt{read\_pages} returns selected pages. By default, a window returns at most 3,000 characters and a page request at most 6,000 characters; hard caps prevent a call from exceeding 6,000 and 10,000 characters, respectively. The bounds create a deliberate transition from navigation to evidence: Topic labels and snippets can motivate a reading decision, but substantive claims should be grounded in the returned source text, consistent with the distinction between retrieval cues and evidentiary support in grounded generation systems \citep{lewis_retrieval-augmented_2020,gao_retrieval-augmented_2024}.

Evidence responses include document and page provenance, while all responses include warnings, an estimated response size, and suggested next actions. Three constraints therefore govern the interaction. First, \emph{orientation is not evidence}: Derived topics and snippets route attention, while source passages support claims. Second, \emph{scope is explicit}: Corpus-wide search remains available, but the agent must request it rather than repeatedly searching an undefined space. Third, \emph{source access is bounded and auditable}: The interface limits how much text can be returned at once and preserves the document and location from which it came. These constraints make the agent's search strategy more visible and limit the amount of source material introduced at each step.

\section{Evaluation}
I evaluated FieldGuide by measuring whether tool-structured access to a public-records corpus changed how a GenAI agent performed on investigative document tasks. The evaluation was designed to test three properties that matter for journalistic use: whether the agent could work efficiently through a heterogeneous document collection, whether its answers remained faithful to the underlying documents, and whether it could surface leads that human readers judged to be viable starting points for reporting.

\subsection{Setup}
The evaluation corpus consisted of 18 completed Chicago Police Department public-records requests returned between June 1, 2025 and June 1, 2026, collected from MuckRock and excluding requests fulfilled only with audio or video. In total, the corpus contained 38 documents: 37 PDFs and one image. The documents were deliberately heterogeneous, including incident reports, tactical response reports, training materials, policy documents, email records, and procurement files. This mix creates a heterogeneous exploratory setting resembling public-records reporting, in which a reporter is trying to understand what kinds of materials a public agency has produced and how those materials might connect.

I compared three conditions. In the \emph{raw-file} condition, the agent had access only to the source files in the repository. In the \emph{RAG} condition, the agent accessed a separate LanceDB\footnote{\url{https://github.com/lancedb/lancedb}} hybrid-search index through a single MCP tool. That index reused the same extraction and passage-segmentation pipeline, encoded passages with the \texttt{BAAI/bge-small-en-v1.5} model, and combined keyword and embedding-based retrieval. In the \emph{FieldGuide} condition, the agent had access to FieldGuide's orientation, scoped-search, document-triage, and bounded-reading tools through MCP. Each condition was run in a separate repository with no prior thread context or outside information. All runs used Codex Desktop 0.140.0-alpha.19 with GPT-5.5 and high reasoning effort.

Each condition received the same five prompts, selected to vary the specificity and evidentiary burden of the task. The first prompt asked for a broad description of the corpus and its document types. The second asked about Homeland Security training materials. The third asked the agent to find and summarize a police report about a trespassing incident, with references to specific document locations. The fourth asked for the narrative account and key people in a December 31, 2024 use-of-force report. The fifth asked for five potential reporting leads with concrete details and document references. Together, these prompts cover corpus orientation, topical synthesis, targeted lookup, incident reconstruction, and open-ended lead generation.

I measured performance along three dimensions. For efficiency, I extracted total input and output tokens from transcript metadata and measured wall-clock time from the first user-message event to the final assistant answer in each run. For accuracy, I manually audited the final responses for factual errors, omissions, citation issues, and document-detail issues. Each issue received a category and severity code, ranging from minor to severe. The severity distinctions were adapted from prior work on hallucination in LLM outputs \citep{rawte-etal-2023-troubling,hagar_not_2025}; the category-specific definitions appear in Appendix~\ref{app:rubrics}.

For the lead-generation task, I conducted a blind rating exercise. The 15 generated leads, five from each condition, were shuffled and rated by three independent raters: the author, an active news consumer, and a journalist. Raters scored each lead on two five-point scales. \emph{Quality} captured whether the lead was complete, specific, and actionable. \emph{Newsworthiness} captured whether the lead would likely matter to a general news audience. I report descriptive means across the 15 ratings per condition. I also computed Krippendorff's \(\alpha\) for interval data as an inter-rater reliability check \citep{krippendorff2019content}. Agreement was low for quality (\(\alpha=.38\)) and newsworthiness (\(\alpha=.22\)), so the lead ratings should be read as descriptive evidence about perceived characteristics rather than as a precise ranking.

\subsection{Results}

\begin{table*}[t]
\centering
\caption{Evaluation results by condition. Time and token counts are summed across the five runs. Issue counts report manually coded accuracy issues by severity. Lead ratings are means across five leads per condition, each rated by three raters.}
\label{tab:evaluation-results}
\small
\begin{tabular}{lrrrrrr}
\toprule
Condition & Time (min.) & Tokens (M) & Accuracy issues (minor/moderate/severe) & Total issues & Lead quality & Lead newsworthiness \\
\midrule
Raw-file & 10.1 & 2.12 & 2/1/0 & 3 & 3.60 & 3.40 \\
RAG & 7.6 & 2.49 & 2/0/0 & 2 & 2.40 & 2.13 \\
FieldGuide & 6.0 & 1.54 & 4/0/0 & 4 & 2.73 & 2.80 \\
\bottomrule
\end{tabular}
\end{table*}

Table~\ref{tab:evaluation-results} summarizes aggregate outcomes. Across the five prompts, the FieldGuide runs totaled 6.0 minutes and 1.5 million tokens, compared with 10.1 minutes and 2.1 million tokens for raw-file access and 7.6 minutes and 2.5 million tokens for RAG. FieldGuide therefore required 41\% less elapsed time and 27\% fewer tokens than raw-file access, and 21\% less time and 38\% fewer tokens than RAG. The token differences primarily reflect the contexts constructed during the runs rather than the length of the final answers. Input tokens accounted for 98.7\% of all tokens in the raw-file condition, 99.0\% in the FieldGuide condition, and 99.4\% in the RAG condition.

The aggregate advantage was concentrated in the two tasks that required sustained synthesis (Table~\ref{tab:task-tokens}). FieldGuide used more tokens than raw-file access on the corpus overview, Homeland Security, and trespassing prompts, but substantially fewer on use-of-force reconstruction and lead generation. The largest raw-file run---lead generation---accounted for 39\% of that condition’s total token use, while RAG’s corpus-overview run accounted for 44\% of its total. By contrast, FieldGuide’s largest run accounted for 25\% of its total. FieldGuide therefore did not minimize token use on every task; its main effect was to limit the largest episodes of context accumulation.

The accuracy audit identified three issues in the raw-file responses, two in the RAG responses, and four in the FieldGuide responses. The raw-file condition made two minor document-detail errors in the broad overview and one moderate factual error in lead generation. RAG misclassified the corpus's standalone image as a PDF and cited the wrong page for one substantive claim. FieldGuide misstated the overall page count and one category-level page estimate in its broad overview; in the use-of-force response, it omitted the involved officer's name and the partner's name while otherwise reconstructing the incident correctly. No response contained a severe issue, a fabricated source, or a central unsupported claim. 

\begin{table}[ht]
\centering
\caption{Token use by task and condition. Counts include input and output tokens. Percentages compare FieldGuide with each alternative.}
\label{tab:task-tokens}
\scriptsize
\setlength{\tabcolsep}{3pt}
\begin{tabularx}{\columnwidth}{@{}>{\raggedright\arraybackslash}Xrrrrr@{}}
\toprule
Task & Raw & RAG & FG & vs. raw & vs. RAG \\
\midrule
Overview & 249,955 & 1,088,527 & 389,035 & +56\% & $-$64\% \\
Homeland Security & 342,201 & 303,660 & 382,360 & +12\% & +26\% \\
Trespassing & 206,352 & 215,261 & 221,369 & +7\% & +3\% \\
Use of force & 488,357 & 313,928 & 219,646 & $-$55\% & $-$30\% \\
Leads & 830,132 & 568,479 & 326,386 & $-$61\% & $-$43\% \\
\midrule
\textbf{Total} & \textbf{2,116,997} & \textbf{2,489,855} & \textbf{1,538,796} & \textbf{$-$27\%} & \textbf{$-$38\%} \\
\bottomrule
\end{tabularx}
\end{table}

The raw-file leads received the highest observed means for quality and newsworthiness (3.60 and 3.40), followed by FieldGuide (2.73 and 2.80) and RAG (2.40 and 2.13). Each condition produced only five leads, however, and inter-rater agreement was low, particularly for newsworthiness. The ratings should consequently be treated as exploratory evidence about perceived lead viability, not as a stable ranking of the three access strategies.

Taken together, in this evaluation, FieldGuide reduced total runtime token use and elapsed time, largely by avoiding the most token-intensive runs observed under raw-file and RAG access. It did so without an observed severe accuracy failure. However, it was not the lowest-token condition on every task, and the lead ratings leave open whether narrower access can limit editorial discovery.

\section{Discussion}
This study reframes token efficiency as a problem of access design. Much work on efficient LLM use treats the context window as the main site of intervention: Systems retrieve, compress, or summarize text so that a model can answer a question with fewer tokens \citep{lewis_retrieval-augmented_2020,gao_retrieval-augmented_2024,li_compressing_2023,jiang_llmlingua_2023,jiang_longllmlingua_2024}. FieldGuide intervenes earlier in the workflow. Rather than first asking which passages should be placed in context, it asks what an agent needs to know about a corpus before deciding where to search or read. In doing so, it adapts a longstanding progressive disclosure workflow from exploratory interfaces \citep{shneiderman_eyes_1996} to an agent-facing setting. The efficiency gains observed here come from changing the order of operations by which the agent encounters the corpus.

That distinction is particularly relevant for journalism because document exploration often begins before the reporter can formulate a precise information need. RAG-style systems are well suited to grounding answers once a query has been articulated, but investigative work often involves an earlier orientation stage of discovering what the relevant questions might be. This is closer to berrypicking and analytic sensemaking, where search, reading, comparison, and reformulation unfold together \citep{bates_design_1989,pirolli_sensemaking_2005}. FieldGuide's topic hierarchy and document cards provide a compact representation of the collection that supports an early exploratory stage.

The lead-generation results demonstrate the limits of that intervention. FieldGuide preserved factual performance and received higher ratings than the RAG condition on perceived lead quality and newsworthiness, but it did not match the raw-file condition. That pattern suggests that selective access can reduce unnecessary reading while also narrowing opportunities for making editorial connections. For journalism, this is a substantive design tradeoff. A system that directs an agent toward the most relevant topical region may be less likely to expose the agent to weak signals, anomalies, or cross-topic juxtapositions that make a lead interesting. Prior journalism-facing document tools such as Overview and New/s/leak treated corpus structure as something that helps reporters see both dominant patterns and potentially surprising documents \citep{brehmer_overview_2014,yimam_newsleak_2016,wiedemann_newsleak_2018}. Agent-facing orientation layers need a comparable account of breadth.

This points to a more specific design principle: Exploratory access should make coverage visible and consequential. FieldGuide already exposes neighboring topics, diverse examples, and lower-similarity documents, but these remain optional cues. The lead-generation result indicates that optional breadth may be insufficient for open-ended tasks. An exploratory workflow could instead track which branches and documents the agent has inspected, identify unvisited regions, require comparison across distant topics before lead selection, and reserve part of the token budget for anomalous records.

The accuracy audit also separates failures of access from failures of answer construction. FieldGuide’s overview errors concerned corpus-level counts, where derived navigation objects were treated as authoritative metadata. Its use-of-force omissions occurred after the relevant document and incident had been identified, indicating a completeness failure rather than a retrieval failure. These problems call for different safeguards. Corpus statistics should be computed directly and clearly distinguished from derived topic descriptions, while structured factual prompts should trigger a final completeness check against the fields requested by the user. Treating both problems simply as hallucination would obscure where in the workflow the failure occurred \citep{hagar_not_2025}.

\paragraph{Limitations.}
This evaluation estimates workflow-level performance rather than isolating the effect of individual FieldGuide components. Future work could examine the effect of surfacing individual tools to the agent at a more granular level. Second, the evaluation was run with one model, client configuration, and set of prompts. The results demonstrated here may therefore not transfer to other agents or prompting strategies. Finally, the corpus comprised 38 Chicago Police Department documents and may not represent larger, noisier, multilingual, or less structured collections. Future work should use repeated runs across varied corpora.

\paragraph{Conclusion.}
This paper argues that agentic document exploration is not only a retrieval problem, but an orientation problem. FieldGuide suggests that a reusable corpus map can guide an agent through a collection before it begins close reading, reducing the cost of exploratory work without requiring the agent to ingest the full corpus. At the same time, the lead-generation results suggest that efficiency can come at an editorial cost. The design challenge for journalism-facing AI systems is therefore not simply to retrieve less, but to help agents allocate attention in ways that preserve both factual grounding and the possibility of discovery.

\begin{acks}
The author thanks Lily Alexandroff and Nicolás Rivero for their contributions to the tool evaluation and feedback on this draft.
\end{acks}

\bibliographystyle{ACM-Reference-Format}
\bibliography{sample-base}

\appendix
\section{Accuracy audit codebook}
\label{app:rubrics}
Each final response was compared with the source documents. Each distinct issue received one category code and the corresponding severity level shown in Table~\ref{tab:accuracy-rubric}. The totals report distinct coded issues rather than the number of affected responses; a single response could therefore contribute more than one issue.

\begin{table}[H]
\centering
\caption{Accuracy audit codebook.}
\label{tab:accuracy-rubric}
\scriptsize
\setlength{\tabcolsep}{3pt}
\renewcommand{\arraystretch}{0.92}
\begin{tabularx}{\columnwidth}{@{}llX@{}}
\toprule
Code & Severity & Definition \\
\midrule
F3 & Severe & Fabricated event, person, or claim. \\
F2 & Moderate & Fabricated or incorrect detail that changes the substance of an otherwise correct response. \\
F1 & Minor & An incorrect peripheral detail that does not alter the answer's central meaning. \\
O3 & Severe & Key documents or passages that contradict the substance of the claim are omitted. \\
O2 & Moderate & Documents or passages that complicate or extend the substance of the claim are omitted. \\
O1 & Minor & A detail requested by the prompt is omitted, but the answer retains its central substance. \\
C3 & Severe & A claim is unsupported or attributed to a fabricated document. \\
C2 & Moderate & A claim is attributed to the wrong document. \\
C1 & Minor & A claim is attributed to the wrong page or pages in the correct document. \\
D3 & Severe & Nonexistent documents or document topics are referenced. \\
D2 & Moderate & Existing document topics or contents are summarized inaccurately. \\
D1 & Minor & Peripheral document metadata are reported incorrectly without materially changing the answer's characterization of the corpus. \\
\bottomrule
\end{tabularx}
\end{table}

\begin{table}[H]
\centering
\caption{Observed accuracy issues by condition and task. Counts in parentheses indicate multiple issues of the same code.}
\label{tab:accuracy-issues}
\scriptsize
\setlength{\tabcolsep}{3pt}
\renewcommand{\arraystretch}{0.92}
\begin{tabularx}{\columnwidth}{@{}llX@{}}
\toprule
Condition/task & Code & Observed issue \\
\midrule
Raw-file overview & D1 (2) & Incorrect request-folder count and total page count. \\
Raw-file leads & F2 & A partial ammunition figure was treated as representative of the total, substantially understating the number of rounds described in the procurement records. \\
RAG overview & D1 & The standalone image was classified as a PDF. \\
RAG leads & C1 & The firearms-target invoice was cited to the wrong page of the correct document. \\
FieldGuide overview & D1 (2) & Incorrect overall page count and one category-level page estimate. \\
FieldGuide use of force & O1 (2) & Omitted the involved officer's name and the partner's name while retaining other identifying and narrative details. \\
\bottomrule
\end{tabularx}
\end{table}

\end{document}